\documentclass[10pt,twocolumn,letterpaper]{article}

\usepackage[pagenumbers]{cvpr} % To force page numbers, e.g. for an arXiv version

\usepackage[table]{xcolor}
\usepackage{colortbl}
\usepackage{booktabs}
\usepackage{multirow}
\usepackage{array}
\usepackage{makecell}
\usepackage{pifont}
\usepackage{graphicx}
\usepackage{subcaption}
\usepackage{placeins}
\usepackage{afterpage}
\usepackage{stfloats}

\definecolor{cvprblue}{rgb}{0.21,0.49,0.74}
\usepackage[breaklinks,colorlinks,allcolors=cvprblue]{hyperref}

\def\paperID{44068} % *** Enter the Paper ID here
\def\confName{CVPR}
\def\confYear{2026}

\title{Detecting AI-Generated Forgeries via Iterative Manifold Deviation Amplification}

\author{
Jiangling Zhang$^{1}$ \quad
Shuxuan Gao$^{1}$ \quad
Bofan Liu$^{1}$ \quad
Siqiang Feng$^{1}$ \\
Jirui Huang$^{1}$ \quad
Yaxiong Chen$^{1}$\thanks{Corresponding author: yaxiong.chen@whut.edu.cn} \quad
Ziyu Chen$^{1}$ \\
$^{1}$Wuhan University of Technology
}

\begin{document}
\maketitle

\begin{abstract}
The proliferation of highly realistic AI-generated images poses critical challenges for digital forensics, demanding precise pixel-level localization of manipulated regions. Existing methods predominantly learn discriminative patterns of specific forgeries, struggling with novel manipulations as editing techniques evolve. We propose the \textbf{I}terative \textbf{F}orgery \textbf{A}mplifier \textbf{Net}work (\textbf{IFA-Net}), which shifts from learning ``what is fake'' to modeling ``what is real''. Grounded in the principle that all manipulations deviate from the natural image manifold, IFA-Net leverages a frozen Masked Autoencoder (MAE) pretrained on real images as a universal realness prior. Our framework operates through a two-stage closed-loop process: an initial \textbf{D}ual-\textbf{S}tream \textbf{S}egmentation \textbf{N}etwork (\textbf{DSSN}) fuses the original image with MAE reconstruction residuals for coarse localization, then a \textbf{T}ask-\textbf{A}daptive \textbf{P}rior \textbf{I}njection (\textbf{TAPI}) module converts this coarse prediction into guiding prompts that steer the MAE decoder to amplify reconstruction failures in suspicious regions, enabling precise refinement. Extensive experiments on four diffusion-based inpainting benchmarks show that IFA-Net achieves an average improvement of 6.5\% in IoU and 8.1\% in F1-score over the second-best method, while demonstrating strong generalization to traditional manipulation types.
\end{abstract}    

\raggedbottom
\section{Introduction}

\noindent
The rapid evolution of generative models---particularly the rise of diffusion-based approaches~\cite{rombach2022latent,NicholDRSMMSC22,podell2024sdxl,saharia2022imagen}---has fundamentally reshaped the landscape of image synthesis. Modern diffusion models can now produce visuals that are nearly indistinguishable from genuine photographs, blurring the boundary between creation and deception. While these advances enable unprecedented creative applications, they also introduce critical challenges for digital forensics. As manipulated images proliferate across online platforms, reliable pixel-level localization becomes essential. Beyond simply verifying authenticity, localization provides precise regions of manipulation, offering interpretable and actionable forensic evidence.

\begin{figure}[t]
\centering
\includegraphics[width=\linewidth]{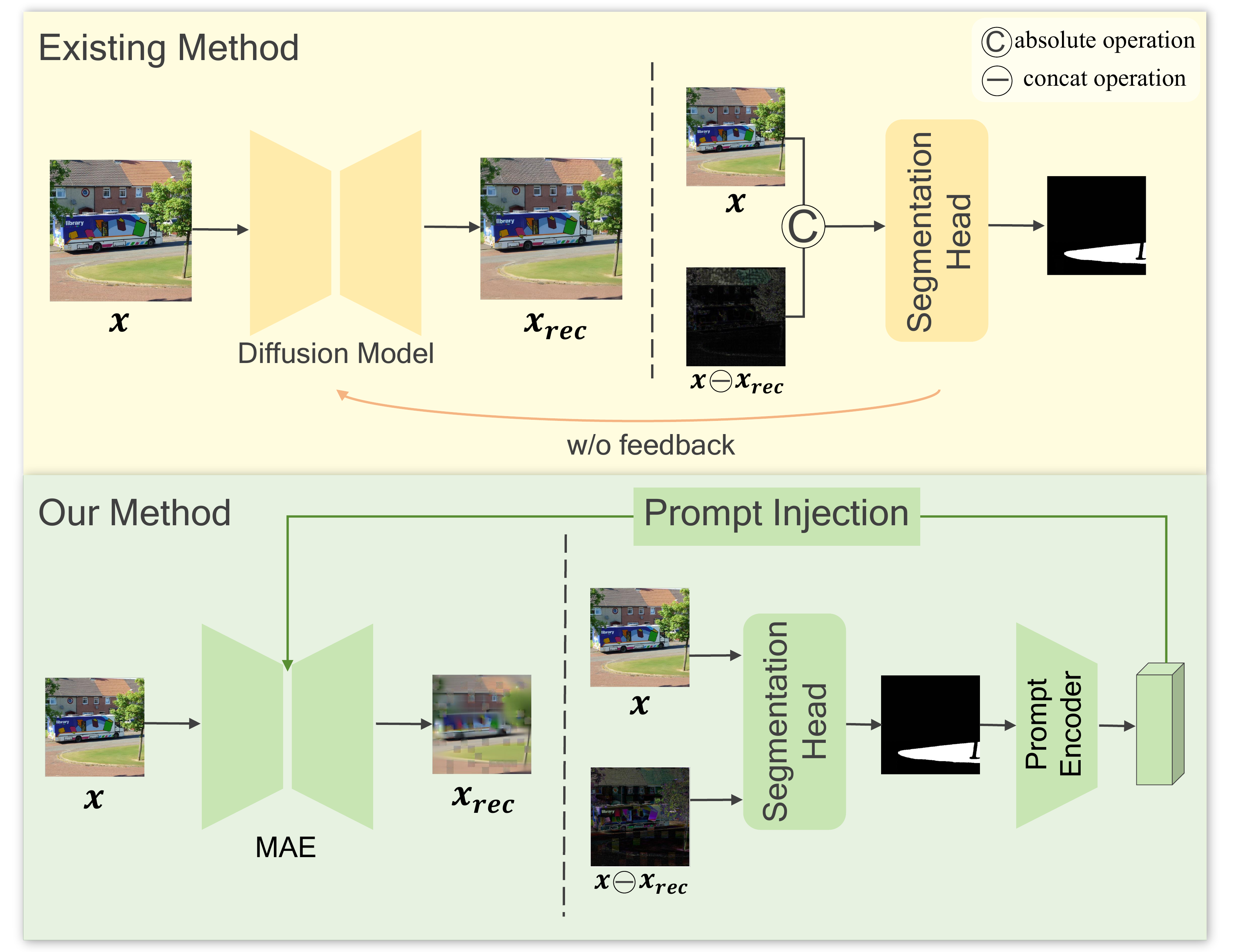} 
\caption{
Conceptual comparison of our IFA-Net against existing methods.
Existing methods often leverage Diffusion Models for reconstruction in a single-pass, open-loop process that lacks task-adaptive guidance.Our method is grounded in a different prior: a Masked Autoencoder (MAE) pretrained on real images. We introduce a two-stage, closed-loop framework where the coarse prediction from Stage 1 guides a second, targeted reconstruction in Stage 2 for significant anomaly amplification.
}
\label{fig:framework_overview}
\end{figure}

\noindent
Despite significant advances, existing research still faces substantial limitations. Early forensic methods that relied on handcrafted cues—such as sensor noise or frequency inconsistencies~\cite{cozzolino2020noiseprint}—collapse when confronted with modern generative pipelines, including subject-driven editing paradigms like DreamBooth~\cite{ruiz2023dreambooth}. Deep learning approaches improve robustness but often at the cost of architectural complexity. For instance, TruFor~\cite{GuillaroCSDV23} integrates RGB and noise residuals through a transformer to detect anomalies;
HiFi-Net~\cite{guo2023hierarchical} employs a multi-branch structure for fine-grained classification and localization;
DiffForensics~\cite{yu2024diffforensics} leverages diffusion priors in a two-stage design;
and UnionFormer~\cite{li2024unionformer} fuses object-level context with multi-view representations.
Although these architectures achieve strong performance, they remain heavy and tightly coupled to specific training distributions, resulting in poor generalization to unseen diffusion models. This persistent tension between specialization and generalization forms a central bottleneck in modern forensic localization. Methods such as MVSS-Net~\cite{dong2022023mvssnet} further highlight this trend, excelling in known domains but struggling with unseen manipulation types.Fundamentally, these approaches learn discriminative patterns of known forgeries rather than modeling the intrinsic properties of authentic images, limiting their ability to detect novel manipulations.

\vspace{4pt}
\noindent
To address these challenges, we introduce IFA-Net, a simple yet powerful framework grounded in the assumption that even the most realistic forgeries inevitably deviate from the natural image manifold. Rather than relying on a monolithic detector, Iterative Forgery Amplifier Network (IFA-Net) decomposes the problem into two complementary stages. In the first stage, a frozen Masked Autoencoder (MAE) pretrained on large-scale natural images acts as a universal artifact amplifier. Its reconstruction process naturally magnifies subtle inconsistencies in manipulated regions, transforming weak and heterogeneous traces into stronger and more coherent signals. In the second stage, a lightweight dual-stream segmentation network jointly analyzes the original and amplified representations. These two streams are bridged through a task-adaptive prior injection (TAPI) module, which dynamically injects the amplified priors into the segmentation process, enabling precise localization with minimal computational overhead.

\vspace{4pt}
\noindent
Extensive experiments on multiple diffusion-based forgery benchmarks demonstrate that IFA-Net achieves consistent state-of-the-art localization accuracy and robust generalization to unseen generators. Our findings reveal that coupling a frozen universal prior with task-adaptive injection effectively bridges the long-standing gap between simplicity and generalization in image forgery localization.

% \vspace{6pt}
% \noindent
% \textbf{Our main contributions are summarized as follows:}
% \begin{itemize}
% \item Universal artifact amplification: A frozen MAE is used as a plug-and-play artifact amplifier that boosts generalization without retraining for specific manipulations.
% \item Task-adaptive prior injection: A lightweight TAPI module integrates amplified priors into the segmentation network, transferring generic anomaly cues into task-specific localization.
% \item Performance with simplicity: IFA-Net achieves state-of-the-art results across diffusion forgery benchmarks while maintaining a compact, efficient, and generalizable architecture.
% \end{itemize}

\vspace{6pt}
\noindent
Our main contributions are summarized as follows:
\begin{itemize}
\item Realness-driven detection paradigm: We model authenticity rather than memorizing forgery patterns, using a frozen MAE pretrained on natural images as a universal realness prior to detect manipulations through their deviations from the natural image manifold.

\item Closed-loop amplification framework: We introduce a two-stage architecture where the Task-Adaptive Prior Injection (TAPI) module creates a feedback loop, converting coarse predictions into prompts that guide reconstruction to progressively amplify weak forgery signals.

\item State-of-the-art with strong generalization: IFA-Net achieves superior performance on diffusion-based benchmarks, and demonstrates robust cross-dataset generalization to unseen generators and traditional tampering methods.
\end{itemize}

\section{Related Work}
\label{sec:related}

\subsection{Generated Image Detection}
With the rapid advancement of diffusion models, synthetic images have become nearly indistinguishable from real ones, posing unprecedented challenges to visual forensics. This evolution marks a paradigm shift from artifact-based discrimination toward generative-aware forensic modeling. Early studies primarily focused on local artifact identification, while recent works have emphasized modeling global consistency and semantic coherence. TruFor~\cite{GuillaroCSDV23} fused RGB texture and noise residuals to improve pixel-level localization, and watermark-based localization approaches such as EditGuard~\cite{ZhangLYXLZ24} further explored structural cues for tamper analysis.

\begin{figure*}[t]
  \centering
  \includegraphics[width=\linewidth]{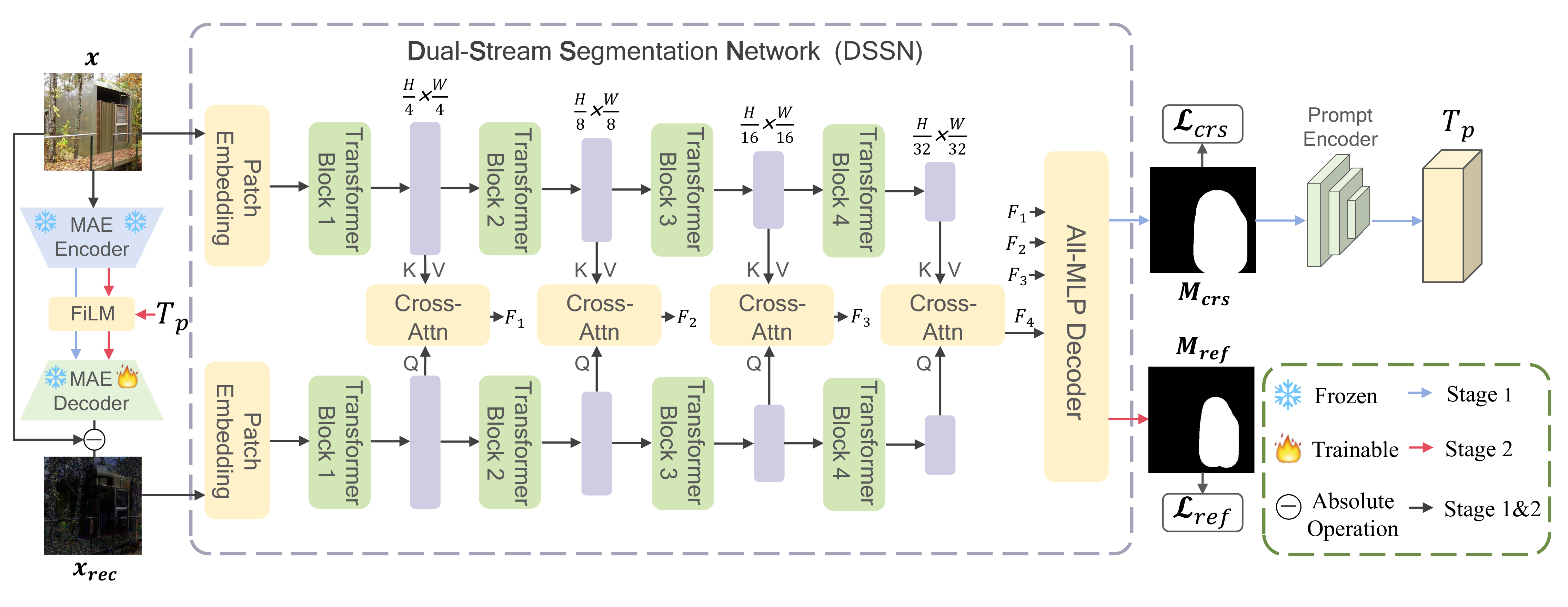}
  \caption{
    Overall architecture of IFA-Net.
    The framework operates in two stages while sharing the same Dual-Stream Segmentation Network (DSSN).
    Stage 1 (blue arrows): The input image $x$ is processed by a fully frozen MAE (Encoder \& Decoder) to produce an initial residual map $x_{\text{rec}}$. The DSSN then fuses $x$ and $x_{\text{rec}}$ to predict a coarse mask $M_{\text{crs}}$, supervised by $\mathcal{L}_{\text{crs}}$.
    Stage 2 (red arrows): The coarse mask $M_{\text{crs}}$ is encoded into Task-Adaptive Prompts ($T_p$), which modulate features from the frozen MAE Encoder via a FiLM layer. The modulated features are passed to a trainable MAE Decoder to generate an amplified residual map, which is then used by the shared DSSN to predict the final refined mask $M_{\text{ref}}$, supervised by $\mathcal{L}_{\text{ref}}$.
  }
  \label{fig:framework}
\end{figure*}

In parallel, frequency-domain and multimodal cues have attracted increasing attention. 
Beyond visual cues, Zhou et al.~\cite{zhou2025aigi} explored vision–language alignment to achieve multimodal consistency. 
However, existing detection paradigms remain largely discriminative and data-driven, relying on local texture or frequency biases within training distributions~\cite{zhu2023genimage}.
 Proactive detection frameworks~\cite{asnani2022proactive} highlight the need to anticipate unseen manipulations rather than fitting to known distributions, while MALP~\cite{asnani2023malp} extends this proactive view to localization by leveraging early anomaly cues. Universal detectors~\cite{ojha2023universal} similarly emphasize generator-agnostic modeling, reinforcing the importance of priors that generalize beyond training data. This motivates a shift toward reconstruction-based modeling that leverages the intrinsic discrepancy between real and synthetic image manifolds.

%------------------------------------------------------------------------
\subsection{Detection Based on Reconstruction Error}
These advances highlight the growing importance of generative priors and naturally motivate reconstruction-error-based detection.As diffusion models continue to improve in reconstruction fidelity, researchers have begun exploiting their inherent limitations in reproducing real images. Recent work explores contrastive reconstruction signals to strengthen discrepancy modeling, such as DRCT~\cite{chen2024drct}. Inversion-based reconstruction methods have also shown promise, e.g., FakeInversion~\cite{cazenavette2024fakeinversion}, where inversion trajectories expose generator-specific inconsistencies. Wang et al.~\cite{wang2023dire} introduced a diffusion reconstruction-error framework capturing pixel-level discrepancies, while training-free reconstruction error detectors such as AEROBLADE~\cite{RickerLF24} and frequency-guided reconstruction methods such as FIRE~\cite{ChuXWZYZ25} further demonstrate the generality of reconstruction-asymmetry cues. Building on latent reconstruction analysis, LARE$^2$~\cite{luo2024lare2} demonstrated that latent-space reconstruction biases can further enhance generalizability across unseen generators. Zhang et al.~\cite{zhang2024imd_inr} proposed an INR-based preprocessing prior that stabilizes reconstruction and suppresses noise-sensitive residuals. Additional decomposition strategies have been discussed in prior work, but such approaches remain limited in their adaptability to diverse manipulation types.

Nevertheless, most reconstruction-based methods employ static generative priors that operate independently of the detection task, lacking adaptive feedback or semantic modulation~\cite{liu2023evp,li2024napl}. As a result, artifact amplification is unguided and reconstruction often emphasizes noise rather than task-relevant cues. To address these limitations, IFA-Net introduces a generative–discriminative collaborative framework that enables the reconstruction prior to be dynamically guided by the detection objective. By treating the pretrained MAE as a universal realness prior, IFA-Net encourages the model to learn which regions conform to the natural image manifold instead of merely memorizing artifact patterns. This design bridges generative modeling and pixel-level localization, achieving a better balance between interpretability, robustness, and generalization.

\raggedbottom
%------------------------------------------------------------------------
\section{Methodology}
\label{sec:methodology}

%------------------------------------------------------------------------
\subsection{Motivation}
\label{sec:motivation}

Image forgery detection faces a fundamental challenge: forged regions are open-world and endlessly diverse, while real images follow a stable and compact distribution. Learning ``what fake looks like'' does not scale with the proliferation of manipulation types. We argue that a more principled strategy is to model \emph{realness} instead of forgery.

A Masked Autoencoder (MAE), pretrained on large-scale natural images, implicitly captures this realness prior~\cite{HeCXLDG22,WangHAZ25}. It reconstructs authentic content with high fidelity but consistently fails on manipulated regions that deviate from the natural image manifold. Such reconstruction asymmetry provides an intrinsic and generalizable signal: if a model knows what is real, whatever it cannot faithfully reconstruct must be suspicious.

However, MAE reconstruction alone yields only coarse discrepancies, offering limited guidance for precise localization. What is missing is a mechanism that can iteratively refine and amplify these weak signals. This motivates a task-adaptive closed loop in which initial anomaly cues guide subsequent reconstruction, enabling the system to progressively highlight regions inconsistent with the realness prior.

Based on this perspective, we design the Iterative Forgery Amplifier Network (IFA-Net) to operationalize a ``Detect--Guide--Amplify'' paradigm: \emph{Detect} deviations from realness, \emph{Guide} the generative prior using task cues, and \emph{Amplify} the anomaly response. This shift from learning forgery patterns to leveraging reconstruction-based realness allows the model to generalize effectively to unseen manipulation types, as it no longer depends on manipulation-specific cues.

A recent work in face forgery detection leverages an end-to-end reconstruction-classification framework~\cite{Cao0YCDY22}, which aligns with our strategy of using reconstruction to identify suspicious regions.

%------------------------------------------------------------------------

\begin{figure}[!t]
  \centering
  \includegraphics[width=\linewidth]{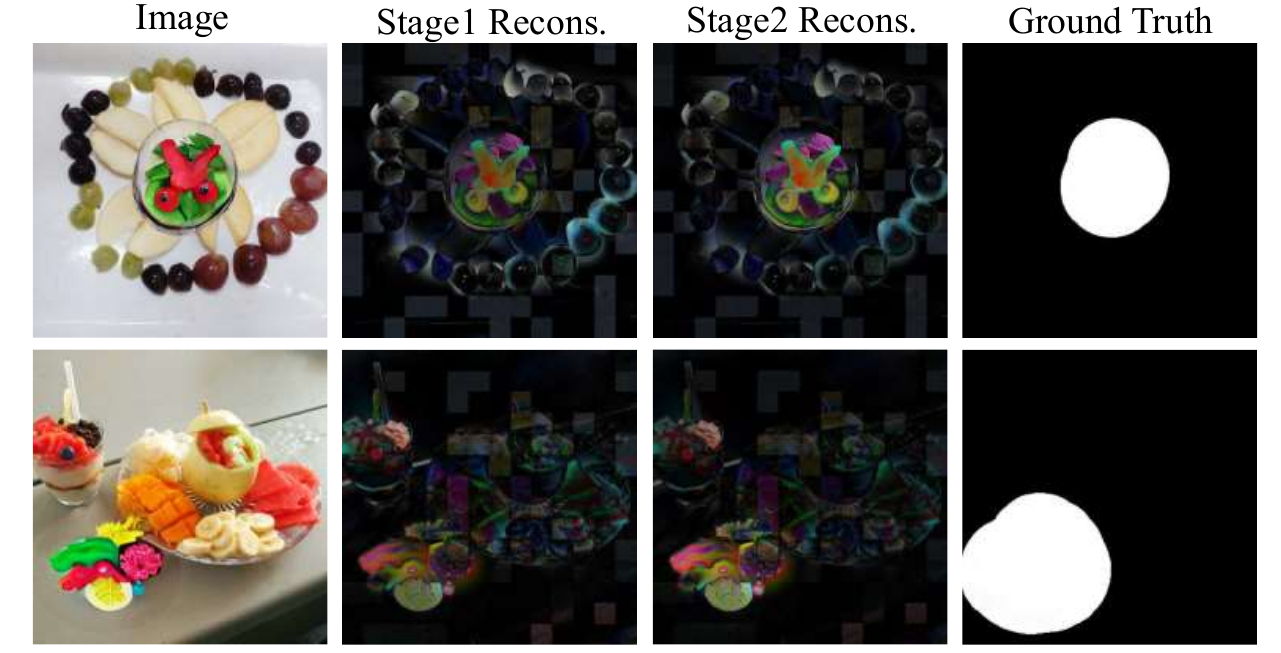}
  \caption{
Visualization of anomaly amplification. We compare reconstructed images and residual maps from Stage 1 and Stage 2. In Stage 1, the frozen MAE can already produce coarse forgery cues—although weak and noisy, they indicate that the MAE indeed encodes a realness prior. In Stage 2, prompt-guided reconstruction forces the model to fail more strongly on forged regions, further amplifying these forgery cues and yielding a much cleaner and more salient residual map that better supports the final segmentation.
  }
  \label{fig:comparison}
\end{figure}

\subsection{Realness-Guided Closed-Loop Detection}
\label{sec:overall_architecture}

As illustrated in Fig.~\ref{fig:framework}, IFA-Net consists of two collaborative stages that share a Dual-Stream Segmentation Network (DSSN), forming a ``Detect--Guide--Amplify'' task-adaptive closed loop grounded in a frozen MAE.

In the first stage, the frozen MAE performs an unconditional reconstruction of the input image and produces an initial reconstruction residual. This residual encodes deviations from the natural image manifold and is fused with the original image by the DSSN to produce a coarse forgery mask. This stage thus focuses on \emph{Anomaly Discovery}, exposing subtle manifold discrepancies without task-specific guidance.

In the second stage, the coarse mask is treated as an explicit structural prior and is converted into task-dependent prompts via a lightweight Prompt Encoder. These prompts are injected into the frozen MAE encoder through a FiLM-based modulation layer inside the proposed Task-Adaptive Prior Injection (TAPI) module. The modulated encoder features are then fed into a trainable decoder, which is encouraged to \emph{fail more strongly} on suspicious regions during reconstruction. The resulting residual, being significantly amplified in forged areas and suppressed in authentic regions, is again processed by the shared DSSN to refine the segmentation mask.

In this way, Stage~I performs anomaly discovery using a universal reconstruction prior, while Stage~II performs guided anomaly amplification by steering the generative prior toward suspicious regions. The two stages are jointly optimized, forming an explicit closed loop between generative modeling (realness prior) and discriminative segmentation (pixel-level localization), and yielding interpretable and robust forgery localization.In recent works, visual prompting has been used to guide segmentation tasks, such as in explicit visual prompting~\cite{LiuSC23} for foreground segmentation.

%------------------------------------------------------------------------

\subsection{Dual-Stream Segmentation Network (DSSN)}
\label{sec:dssn}

We now detail the Dual-Stream Segmentation Network (DSSN), which is shared across both stages. The DSSN is designed to jointly capture high-level semantic structure and low-level artifact cues, enabling precise discrimination between authentic and manipulated regions.

Compared with the original SegFormer~\cite{XieWYAAL21}, DSSN incorporates dual inputs (the original image and an MAE residual map) and introduces cross-stream fusion atop dual encoders. This design is inspired by multi-source manipulation localization networks such as CFL-Net~\cite{NiloyBW23} and MFINet~\cite{RenHNXZW24}, which demonstrate the benefit of contrastive multi-branch cues.

%------------------------------------------------------------------------
\subsubsection{Encoder Structure}

The input image $\mathbf{X} \in \mathbb{R}^{H \times W \times C}$ and a residual map $\mathbf{X}_\text{rec} \in \mathbb{R}^{H \times W \times C}$ are fed into a content stream and an artifact stream, respectively, where $H$ and $W$ denote the spatial resolution and $C$ denotes the number of channels. The content stream focuses on high-level semantics and scene structure, while the artifact stream focuses on fine-grained inconsistencies highlighted by the MAE residual.

Each stream applies an overlapped patch embedding module followed by a hierarchical transformer backbone, producing multi-scale features at a fixed number of stages. The content and artifact features at stage~$\ell$ are denoted as $\mathbf{F}^{\ell}_\text{con} \in \mathbb{R}^{B \times N_\ell \times D_\ell}$ and $\mathbf{F}^{\ell}_\text{art} \in \mathbb{R}^{B \times N_\ell \times D_\ell}$, where $B$ is the batch size, $N_\ell$ is the number of tokens and $D_\ell$ is the channel dimension at stage~$\ell$.

To enable interaction between semantic and artifact information, we apply cross-stream fusion via multi-head cross-attention at each stage. Specifically, the content features act as queries and the artifact features act as keys and values:
\begin{equation}
\tilde{\mathbf{F}}^{\ell}_\text{con} = \text{CrossAttn}(\mathbf{F}^{\ell}_\text{con}, \mathbf{F}^{\ell}_\text{art}, \mathbf{F}^{\ell}_\text{art}),
\end{equation}
where $\text{CrossAttn}(\mathbf{Q}, \mathbf{K}, \mathbf{V})$ denotes multi-head cross-attention with queries $\mathbf{Q}$, keys $\mathbf{K}$, and values $\mathbf{V}$. This design allows semantic tokens to attend to artifact cues, progressively integrating manifold discrepancy information into the semantic representation. The fused content features at each stage, denoted by $\tilde{\mathbf{F}}^{\ell}_\text{con}$, form a complementary multi-scale representation that captures both global context and local inconsistencies.

%------------------------------------------------------------------------
\subsubsection{Decoder and Output}

The DSSN decoder aggregates the multi-scale fused features $\tilde{\mathbf{F}}^{\ell}_\text{con}$ into a dense prediction map. Concretely, all stages are first projected to a unified channel dimension and then upsampled to a common spatial resolution. These feature maps are concatenated along the channel dimension, compressed via a linear projection, and finally upsampled to the original image resolution to produce a single-channel forgery probability map. A Sigmoid activation is applied to convert this map into a pixel-wise probability mask.

The same decoder (and encoder weights) are shared between Stage~I and Stage~II. This weight sharing enforces consistent feature usage across coarse and refined predictions, reduces the parameter count, and stabilizes the closed-loop training process.

%------------------------------------------------------------------------
\subsection{Task-Adaptive Prior Injection (TAPI)}
\label{subsec:tapi}

The core mechanism of IFA-Net lies in constructing a two-stage closed loop based on the MAE realness prior~\cite{HeCXLDG22}, dynamically coupling generative reconstruction and discriminative segmentation. The proposed Task-Adaptive Prior Injection (TAPI) module implements the ``Guide--Amplify'' steps by converting coarse masks into prompts and using them to modulate MAE encoder features.

The frozen MAE encoder $E_\text{MAE}(\cdot)$ and the trainable decoder used in Stage~II $D_\text{MAE}(\cdot)$ form the generative backbone. Given an input image $\mathbf{X}$, the encoder produces a token sequence $\mathbf{Z} \in \mathbb{R}^{B \times N \times D}$, where $N$ is the number of patches and $D$ is the token dimension. TAPI uses the coarse mask from Stage~I to generate task-dependent prompts and then modulates $\mathbf{Z}$ via FiLM to guide the subsequent reconstruction.

%------------------------------------------------------------------------
\subsubsection{Prompt Encoding}
\label{subsubsec:prompt_encoding}

The coarse mask predicted by Stage~I, denoted as $\mathbf{M}_\text{crs} \in \mathbb{R}^{B \times H \times W}$, is converted by the Prompt Encoder into a compact set of task-adaptive tokens $\mathbf{T} \in \mathbb{R}^{B \times N_p \times D_p}$, where $N_{p}$ is the number of prompt tokens and $D_{p}$ is the prompt embedding dimension.

Internally, the Prompt Encoder first reduces the spatial resolution of $\mathbf{M}_\text{crs}$ through a stack of convolutional layers with downsampling, producing a feature map of size $B \times C_p \times H_p \times W_p$, where $H_p$ and $W_p$ are lower than $H$ and $W$ by a fixed factor and $C_{p}$ denotes the intermediate channel dimension. This feature map is then reshaped into $B$ sequences of length $N_{p} = H_p W_p$ and linearly projected to dimension $D_{p}$, yielding the prompt token matrix $\mathbf{T}$. These tokens summarize the spatial distribution and shape of suspicious regions and will be used to generate FiLM parameters.

%------------------------------------------------------------------------
\subsubsection{Guided Reconstruction via FiLM}
\label{subsubsec:film}

To inject task information into the MAE encoder features, we adopt Feature-wise Linear Modulation (FiLM). From the prompt tokens $\mathbf{T}$, we first obtain a global summary vector by applying average pooling along the token dimension. Two lightweight parametric mappings then transform this summary vector into scaling and shifting coefficients $\boldsymbol{\gamma} \in \mathbb{R}^{B \times D}$ and $\boldsymbol{\beta} \in \mathbb{R}^{B \times D}$, respectively.
These coefficients are broadcast along the token dimension and applied to the encoder tokens $\mathbf{Z}$ in a feature-wise affine transformation:
\begin{equation}
\tilde{\mathbf{Z}} = \boldsymbol{\gamma} \odot \mathbf{Z} + \boldsymbol{\beta},
\end{equation}
where $\odot$ denotes element-wise multiplication with broadcasting.

The modulated encoder tokens $\tilde{\mathbf{Z}}$ are passed to the trainable decoder $D_\text{MAE}(\cdot)$, which produces a guided reconstructed image and its corresponding residual. Importantly, the MAE encoder $E_\text{MAE}(\cdot)$ remains frozen in both stages, preserving a stable realness prior, while only $D_\text{MAE}(\cdot)$ is fine-tuned in Stage~II to respond to the FiLM modulation. This design encourages the decoder to systematically alter its reconstruction strategy in regions flagged by the prompts, effectively forcing stronger reconstruction failures on forged areas and yielding an amplified, more structured residual map (see Fig.~\ref{fig:comparison}).

%------------------------------------------------------------------------
\subsection{Loss Function}
\label{sec:loss}

IFA-Net jointly optimizes two segmentation branches in an end-to-end manner to ensure stable coarse localization and accurate refined detection. The input image is denoted as $\mathbf{X}$ and the corresponding ground-truth mask is denoted as $\mathbf{M}^*$. The loss of the first stage is defined as the combination of Binary Cross-Entropy (BCE) and Dice loss~\cite{RonnebergerFB15}:
\begin{equation}
\mathcal{L}_\text{crs} = \mathcal{L}_\text{BCE}(\mathbf{M}_\text{crs}, \mathbf{M}^*) + \mathcal{L}_\text{Dice}(\mathbf{M}_\text{crs}, \mathbf{M}^*),
\end{equation}
where $\mathbf{M}_\text{crs}$ denotes the coarse mask predicted by Stage~I. This term enforces reliable anomaly discovery and provides a stable guidance signal for the following refinement stage.

The refinement-stage loss for Stage~II is formulated in the same way:
\begin{equation}
\mathcal{L}_\text{ref} = \mathcal{L}_\text{BCE}(\mathbf{M}_\text{ref}, \mathbf{M}^*) + \mathcal{L}_\text{Dice}(\mathbf{M}_\text{ref}, \mathbf{M}^*),
\end{equation}
where $\mathbf{M}_\text{ref}$ is the refined mask obtained after prior injection and anomaly amplification. It encourages precise pixel-level localization and sharper boundaries.

Finally, the overall training objective combines the two losses as
\begin{equation}
\mathcal{L}_\text{total} = \mathcal{L}_\text{ref} + \alpha \mathcal{L}_\text{crs},
\end{equation}
where the balancing coefficient $\alpha$ is set to $0.5$, which controls the contribution of the coarse-stage supervision. A moderate weight stabilizes the training process while ensuring that the refined prediction dominates the final optimization.        

\section{Experiments and Results}
\label{sec:experiments}
%------------------------------------------------------------------------

\begin{table*}[t]
\centering
\caption{Quantitative results on GIT and TT benchmarks.
Comparison of the proposed method with recent state-of-the-art approaches.
Values are reported as IoU / F1. 
Best and second-best results for IoU and F1 are independently shown in \textbf{bold} and \underline{underline}, respectively.}
\resizebox{\textwidth}{!}{
\begin{tabular}{ccccccccccccccccc}
\hline
&
\multicolumn{2}{c}{\textbf{TruFor}} &
\multicolumn{2}{c}{\textbf{PSCC-Net}} &
\multicolumn{2}{c}{\textbf{MVSS-Net}} &
\multicolumn{2}{c}{\textbf{SPAN}} &
\multicolumn{2}{c}{\textbf{IML-ViT}} &
\multicolumn{2}{c}{\textbf{MaskCLIP}} &
\multicolumn{2}{c}{\textbf{DcDsDiff}} &
\multicolumn{2}{c}{\textbf{Ours}} \\
\textbf{Dataset} & IoU & F1 & IoU & F1 & IoU & F1 & IoU & F1 & IoU & F1 & IoU & F1 & IoU & F1 & IoU & F1 \\
\hline
\rowcolor[HTML]{F2F2F2}
\multicolumn{17}{c}{\textbf{GIT Benchmarks}} \\
\hline
OpenSDI & 0.369 & 0.426 & 0.297 & 0.368 & 0.298 & 0.349 & 0.284 & 0.349 & 0.325 & 0.373 & \underline{0.427} & \underline{0.494} & 0.385 & 0.445 & \textbf{0.487} & \textbf{0.620} \\
GIT10K  & 0.770 & 0.815 & 0.392 & 0.754 & 0.453 & 0.518 & 0.137 & 0.257 & \underline{0.882} & 0.859 & 0.694 & 0.765 & 0.801 & \underline{0.864} & \textbf{0.928} & \textbf{0.963} \\
CocoGlide & 0.481 & 0.552 & 0.769 & 0.869 & 0.419 & 0.533 & 0.276 & 0.411 & 0.291 & 0.440 & 0.783 & 0.878 & \underline{0.878} & \underline{0.935} & \textbf{0.887} & \textbf{0.940} \\
Inpaint32K & 0.711 & 0.762 & 0.116 & 0.207 & 0.496 & 0.578 & 0.456 & 0.541 & 0.703 & \underline{0.775} & 0.390 & 0.480 & \underline{0.790} & 0.852 & \textbf{0.811} & \textbf{0.896} \\
\hline
AVG & 0.583 & 0.639 & 0.394 & 0.550 & 0.416 & 0.494 & 0.288 & 0.390 & 0.550 & 0.612 & 0.573 & 0.654 & \underline{0.713} & \underline{0.774} & \textbf{0.778} & \textbf{0.855} \\
\hline
\rowcolor[HTML]{F2F2F2}
\multicolumn{17}{c}{\textbf{TT Benchmarks}} \\
\hline
IMD & \textbf{0.552} & \underline{0.601} & 0.276 & 0.437 & 0.331 & 0.393 & 0.153 & 0.264 & \underline{0.548} & \textbf{0.619} & 0.541 & 0.569 & 0.396 & 0.568 & 0.392 & 0.563 \\
NIST16 & 0.447 & 0.522 & 0.196 & 0.294 & 0.181 & 0.226 & 0.396 & 0.582 & 0.242 & 0.326 & 0.694 & 0.751 & \underline{0.843} & \underline{0.915} & \textbf{0.949} & \textbf{0.974} \\
CASIA & \underline{0.637} & \underline{0.724} & 0.295 & 0.442 & 0.284 & 0.427 & 0.137 & 0.170 & \textbf{0.675} & \textbf{0.806} & 0.594 & 0.625 & 0.637 & 0.713 & 0.416 & 0.587 \\
\hline
AVG & 0.545 & 0.616 & 0.256 & 0.391 & 0.265 & 0.349 & 0.229 & 0.338 & 0.488 & 0.584 & \underline{0.610} & 0.648 & \textbf{0.625} & \textbf{0.732} & 0.586 & \underline{0.708} \\
\hline
\end{tabular}
}
\label{tab:git_tt_benchmarks}
\end{table*}

\subsection{Experimental Setup}

\noindent
\textbf{Datasets.} 
We evaluate IFA-Net across seven public benchmarks spanning both diffusion-based and GAN-based Generative Image Tampering (GIT) 
and Traditional Tampering (TT) scenarios. 
To ensure consistent input across all settings, all images are resized to $512\times512$.

These benchmarks collectively cover a wide range of manipulation types. 
For Generative Image Tampering, 
1) \textit{OpenSDID}~\cite{WangHH25} provides diverse inpainting prompts and mask shapes for modeling diffusion-aware inconsistencies; 
2) \textit{GIT10K}~\cite{HaoN0W25} contains 10{,}000 authentic–forged pairs with varying mask scales and editing operations; 
3) \textit{CocoGlide}~\cite{NicholDRSMMSC22,LinMBHPRDZ14} derives high-fidelity semantic inpainting samples from COCO using Glide and Stable Diffusion; 
and 4) \textit{Inpaint32K}~\cite{WangNH025} includes restorations produced by diffusion, GAN, and CNN-based inpainting models with diverse masks, enabling cross-mask generalization testing.
For Traditional Tampering, 
5) \textit{IMD2020}~\cite{NovozamskyMS20} contains 2{,}010 manipulated images covering common splicing and copy–move cases; 
6) \textit{NIST16}~\cite{guan2019mfc} provides high-resolution forgeries for device- and compression-robustness evaluation; 
and 7) \textit{CASIA}~\cite{DongWT13} serves as a classical large-scale benchmark for traditional manipulation localization.

We first pretrain our model on the training split of OpenSDID, which contains images generated by Stable Diffusion~1.5. 
The pretrained weights are then used for dataset-specific fine-tuning and evaluation on other benchmarks. 
This training strategy enables the model to learn general diffusion-aware priors while allowing fair cross-dataset comparison.

\vspace{4pt}
\noindent
\textbf{Baselines.}
We compare against a comprehensive set of state-of-the-art methods covering both traditional tampering localization and diffusion-based editing forensics. 
Conventional baselines include TruFor~\cite{GuillaroCSDV23}, PSCC-Net~\cite{LiuLCL22}, MVSS-Net~\cite{DongCHCL23}, and SPAN~\cite{HuZJCYN20}. 
For generative-editing scenarios, we benchmark against Transformer- and diffusion-driven methods such as IML-ViT~\cite{ma2023imlvit}, 
MaskCLIP~\cite{DongBZZCY00YC0Y23}, and DcDsDiff~\cite{HaoN0W25}. 
This collection spans pixel-level, multi-scale, frequency-guided, and reconstruction-based paradigms.

\vspace{4pt}
\noindent
\textbf{Evaluation Metrics.}
Following common practice, we report \textit{F1-score} and \textit{Intersection-over-Union (IoU)}. 
F1 reflects the trade-off between precision and recall, indicating the overall detection capability, 
while IoU quantifies the spatial overlap between the predicted and ground-truth masks, highlighting localization accuracy.

\vspace{4pt}
\noindent
\textbf{Implementation Details.}
IFA-Net is implemented in PyTorch and trained with Distributed Data Parallel (DDP) on eight NVIDIA RTX 4090 GPUs. 
The MAE backbone uses ViT-Large weights pretrained on ImageNet-1k, and the dual-stream segmentation network adopts MiT-B2 weights pretrained on ADE20K. 
Training is performed for up to 100 epochs with an effective batch size of 32 using AdamW (initial learning rate 2e-5, weight decay 1e-5). 
An early-stopping criterion based on validation IoU (patience 10 epochs) selects the final model.

\begin{figure*}[!t]
\centering
\includegraphics[width=\linewidth]{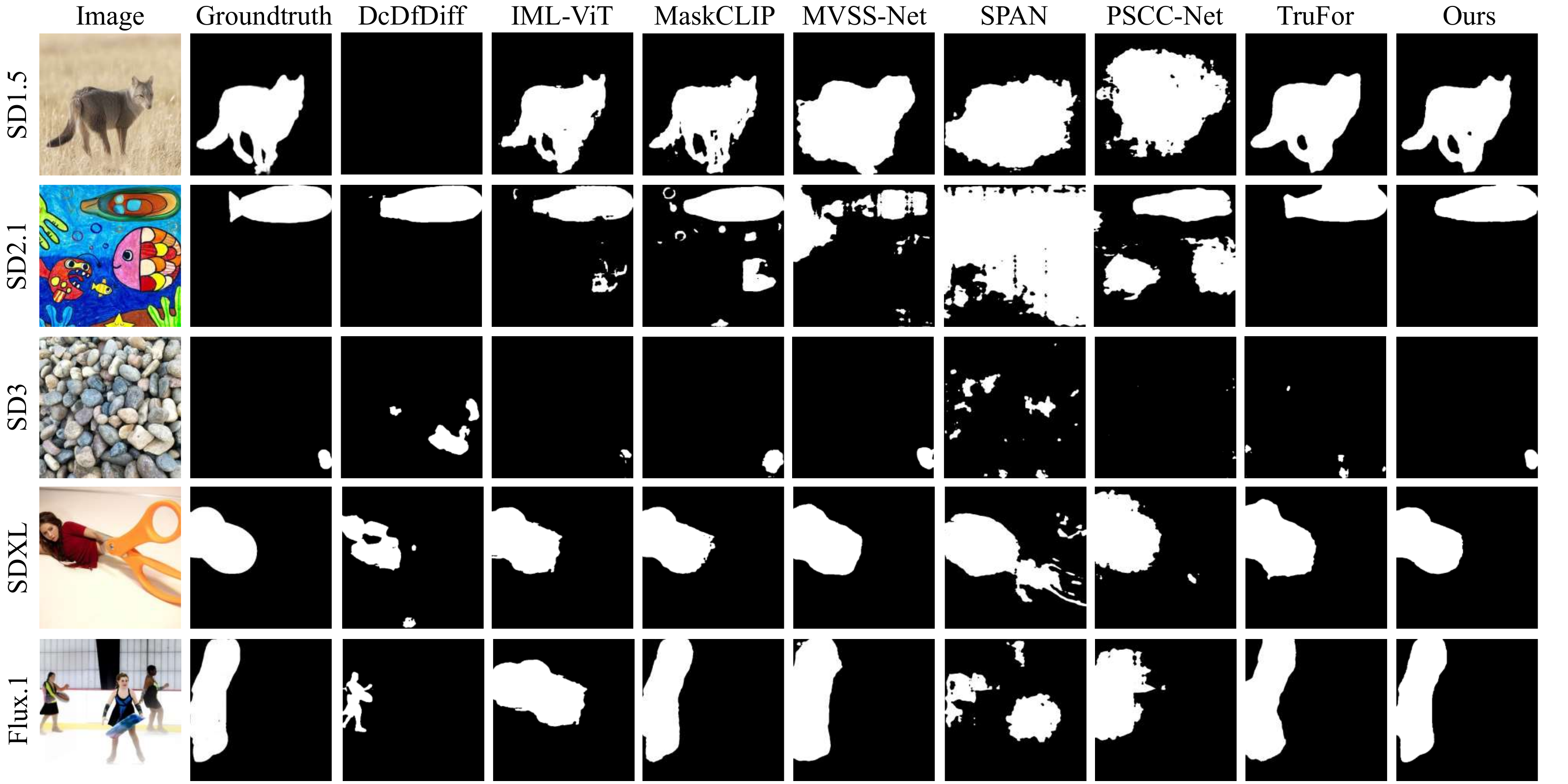}
\caption{Qualitative comparison on OpenSDID.
Examples are selected from five diffusion generators (SD1.5, SD2.1, SD3, SDXL, and Flux.1). 
IFA-Net produces cleaner and more accurate localization masks that align closely with the ground truth, 
whereas other methods often yield incomplete or fragmented detections.}
\label{fig:qualitative_comparison}
\end{figure*}

\subsection{State-of-the-Art Comparison}
%------------------------------------------------------------------------

We now compare our method with recent state-of-the-art approaches on both generative-editing (GIT) and traditional tampering (TT) benchmarks to comprehensively evaluate its effectiveness and generalization.

\vspace{3pt}
\noindent\textbf{Results on GIT Benchmarks.}
Table~\ref{tab:git_tt_benchmarks} compares IFA-Net with existing state-of-the-art detectors across four generative-editing benchmarks—\textit{OpenSDID}, \textit{GIT10K}, \textit{CocoGlide}, and \textit{Inpaint32K}. 
IFA-Net consistently achieves the best or second-best performance on all datasets, reaching an average IoU of 0.778 and F1 of 0.855, establishing a new state of the art for generative forgery localization.

On \textit{OpenSDID}, where our model is pretrained, IFA-Net achieves an IoU of 0.487 and F1 of 0.620, demonstrating the effectiveness of our realness-driven paradigm.
More importantly, the model exhibits strong cross-dataset generalization: on \textit{GIT10K}, \textit{CocoGlide}, and \textit{Inpaint32K}, IFA-Net maintains superior or highly competitive performance, indicating robust transferability across diverse diffusion models, prompt conditions, and mask complexities.
This cross-dataset stability demonstrates that modeling authenticity via a frozen MAE prior—rather than memorizing dataset-specific artifact patterns—enables the network to capture generalizable deviations from the natural image manifold.

%--------------------- Table 3 ---------------------
\begin{table}[t]
\centering
\caption{Ablation study of IFA-Net.
Performance impact of each component on diffusion-based (GIT-AVG) and traditional tampering (TT-AVG) benchmarks.
Index I represents a single-stream baseline that concatenates the original image and residual map as input.
AD denotes the Adaptive Decoder. 
Best results are highlighted in \textbf{bold}.}
\resizebox{\columnwidth}{!}{
\begin{tabular}{cccccccc}
\toprule
& \multicolumn{3}{c}{\textbf{Components}}
& \multicolumn{2}{c}{\textbf{GIT-AVG}}
& \multicolumn{2}{c}{\textbf{TT-AVG}} \\
\cmidrule(lr){2-4} \cmidrule(lr){5-6} \cmidrule(lr){7-8}
\textbf{Index} & \textbf{DSSN} & \textbf{TAPI} & \textbf{AD} & IoU & F1 & IoU & F1 \\
\midrule
I   & \ding{55} & \ding{55} & \ding{55} & 0.670 & 0.770 & 0.459 & 0.608 \\
II  & \ding{51} & \ding{55} & \ding{55} & 0.684 & 0.782 & 0.483 & 0.625 \\
III & \ding{51} & \ding{51} & \ding{55} & 0.745 & 0.823 & 0.554 & 0.670 \\
IV  & \ding{51} & \ding{51} & \ding{51} & \textbf{0.778} & \textbf{0.855} & \textbf{0.586} & \textbf{0.708} \\
\bottomrule
\end{tabular}}
\label{tab:ablation}
\end{table}

Beyond quantitative metrics, our approach also improves interpretability.
By grounding detection in the reconstruction behavior of a real-image prior, IFA-Net provides explicit visual cues corresponding to physical inconsistencies, making predicted masks more semantically meaningful for forensic applications.

\vspace{3pt}
\noindent\textbf{Results on TT Benchmarks.}
Although primarily designed for diffusion-based forgeries, IFA-Net generalizes well to conventional tampering datasets including \textit{IMD2020}, \textit{NIST16}, and \textit{CASIA}, which involve copy–move, splicing, and region removal operations.

As shown in Table~\ref{tab:git_tt_benchmarks}, IFA-Net achieves an average F1 of 0.708, comparable to DcDsDiff and outperforming most classical detectors.
This demonstrates that the learned realness prior from large-scale authentic imagery generalizes beyond diffusion or semantic editing, enabling IFA-Net to capture structural and texture inconsistencies inherent to traditional tampering.
In particular, the frozen MAE prior preserves natural statistical dependencies, allowing our model to identify subtle copy–move traces and splicing boundaries that often confuse purely discriminative methods.

\vspace{3pt}
\noindent\textbf{Qualitative Comparisons.}
Figure~\ref{fig:qualitative_comparison} presents qualitative results on the \textit{OpenSDID} dataset.
Across five diffusion generators—SD1.5, SD2.1, SD3, SDXL, and Flux.1—IFA-Net produces cleaner and more coherent localization masks that align closely with the ground truth.
Predicted boundaries are sharper and more complete, showing the model's ability to suppress background noise and to highlight fine-grained diffusion traces or small edited objects.
This qualitative consistency supports our claim that the proposed closed-loop amplification not only enhances numerical performance but also yields visually interpretable evidence.

\subsection{Ablation Study}

To assess the individual and combined contributions of each component in IFA-Net, we conduct a comprehensive ablation study on both diffusion-based (GIT-AVG) and traditional tampering (TT-AVG) benchmarks. Starting from a single-stream baseline, we progressively introduce the Dual-Stream Segmentation Network (DSSN), the Task-Adaptive Prior Injection module (TAPI), and the Adaptive Decoder (AD).

Table 2 presents the quantitative results. The baseline configuration (Index I) achieves 0.670 IoU and 0.770 F1 on GIT-AVG, and 0.459 IoU and 0.608 F1 on TT-AVG. This limited performance stems from the model's inability to separately process semantic content and artifact residuals. Introducing the DSSN (Index II) brings moderate improvements of 1.4\% IoU and 1.2\% F1 on GIT-AVG, and 2.4\% IoU and 1.7\% F1 on TT-AVG, attributed to its dual-encoder architecture that decouples semantic and artifact cues.

Adding the TAPI module (Index III) yields the most substantial boost, with 6.1\% IoU and 4.1\% F1 improvements on GIT-AVG, and 7.1\% IoU and 4.5\% F1 on TT-AVG compared to DSSN alone. This validates that task-adaptive guidance is essential for anomaly amplification. The TAPI module converts coarse predictions into structural prompts and injects them via FiLM-based modulation, steering reconstruction to emphasize manifold deviations. The Adaptive Decoder (Index IV) further refines boundary precision, providing additional gains of 3.3\% IoU and 3.2\% F1 on GIT-AVG, and 3.2\% IoU and 3.8\% F1 on TT-AVG. Overall, the complete IFA-Net surpasses the baseline by 10.8\% IoU and 8.5\% F1 on GIT-AVG, and 12.7\% IoU and 10.0\% F1 on TT-AVG, demonstrating strong synergy among all components.

\subsection{Robustness Evaluation}

\begin{figure}[t]
  \centering
  \includegraphics[width=\linewidth]{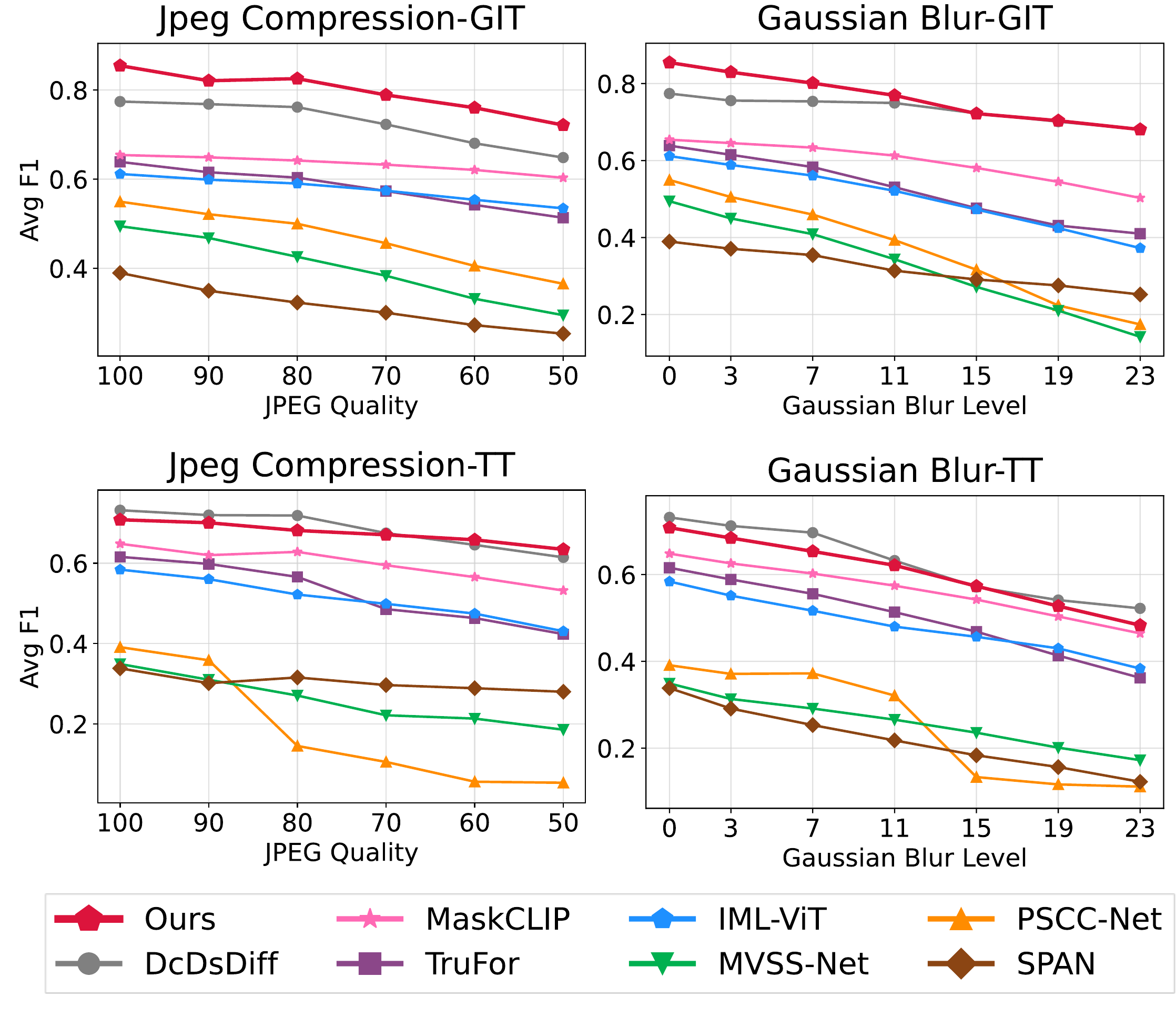}
  \caption{
    Robustness analysis under Gaussian blur and Jpeg compression.
    % IFA-Net maintains the highest F1-scores and exhibits the smoothest degradation curve 
    % across both generative-editing (\textit{GIT}) and traditional tampering (\textit{TT}) domains.
    % These results demonstrate that the closed-loop amplification and task-adaptive prior injection 
    % effectively enhance stability against low-level degradations.
  }
  \label{fig:robustness}
\end{figure}

To assess the stability of IFA-Net under common post-processing distortions, we evaluate its robustness against Jpeg compression and Gaussian blur on both diffusion-based (GIT) and traditional tampering (TT) benchmarks. Figure~\ref{fig:robustness} reports the average F1-scores under increasing distortion levels.

Under Jpeg compression in the GIT domain, IFA-Net maintains the highest F1-scores at high quality levels (100-80), benefiting from task-adaptive prior injection that emphasizes manifold discrepancy rather than compression-sensitive artifacts. As compression quality decreases to 60 and 50, the performance gap narrows with MaskCLIP and DcDsDiff, though IFA-Net remains competitive. For Gaussian blur perturbations in the GIT domain, our method achieves strong F1-scores at lower blur levels (0-11), outperforming most baselines. However, as blur intensity increases beyond level 15, DcDsDiff exhibits better resilience. The diffusion-prior-based reconstruction preserves global structural cues more effectively under severe smoothing conditions.

In the TT domain, IFA-Net outperforms conventional detectors under Jpeg compression at high qualities (100-80), while DcDsDiff shows stronger resilience under severe compression (60-50). The reconstruction-based framework in DcDsDiff maintains better stability when handling quantization artifacts introduced by aggressive compression. For Gaussian blur, IFA-Net shows competitive performance at moderate blur levels (0-11), but DcDsDiff demonstrates superior stability at higher blur levels. The global reconstruction framework in DcDsDiff is less affected by texture smoothing compared to our manifold-based residual amplification approach. Despite these limitations under extreme perturbations, IFA-Net consistently surpasses PSCCNet, MVSS-Net, and SPAN across most distortion levels, validating that the closed-loop amplification provides a resilient foundation for practical forensic scenarios.

%------------------------------------------------------------------------
\section{Conclusion}
%------------------------------------------------------------------------

We presented IFA-Net, a realness-driven framework for forgery localization. 
Instead of learning forgery patterns, IFA-Net models authenticity by exploiting the reconstruction asymmetry of a frozen MAE trained on real images. 
Through a two-stage closed-loop design—anomaly discovery and anomaly amplification—the model converts reconstruction residuals into explicit localization cues. 
This enables the network to learn \textit{what is real} rather than memorize \textit{what is fake}, achieving robust and explainable performance across diverse manipulations. 
Experiments confirm its superiority under perturbations such as compression and blur. 
Future work will extend this paradigm to video and multimodal forensics, and explore weakly supervised optimization toward a unified, realness-centric forensic framework.

{
    \small
    \bibliographystyle{ieeenat_fullname}
    \bibliography{main}

@String(CVPR= {IEEE Conf. Comput. Vis. Pattern Recog.})

@String(ICCV= {Int. Conf. Comput. Vis.})

@String(ECCV= {Eur. Conf. Comput. Vis.})

@String(ICLR = {Int. Conf. Learn. Represent.})

@String(IJCAI = {IJCAI})

@String(AAAI = {AAAI})

@String(CVPR  = {CVPR})

@String(ICCV  = {ICCV})

@String(ECCV  = {ECCV})

@String(ICLR  = {ICLR})

@inproceedings{rombach2022latent,
  title     = {High-Resolution Image Synthesis with Latent Diffusion Models},
  author    = {Rombach, Robin and Blattmann, Andreas and Lorenz, Dominik and Esser, Patrick and Ommer, Björn},
  booktitle = {IEEE/CVF Conference on Computer Vision and Pattern Recognition (CVPR)},
  pages     = {10674--10685},
  year      = {2022},
  doi       = {10.1109/CVPR52688.2022.01042}
}

@inproceedings{podell2024sdxl,
  title     = {SDXL: Improving Latent Diffusion Models for High-Resolution Image Synthesis},
  author    = {Podell, Dustin and English, Zion and Lacey, Kyle and Blattmann, Andreas and Dockhorn, Tim and Müller, Jonas and Penna, Joe and Rombach, Robin},
  booktitle = {International Conference on Learning Representations (ICLR)},
  year      = {2024},
  url       = {https://openreview.net/forum?id=di52zR8xgf}
}

@inproceedings{guo2023hierarchical,
  title     = {Hierarchical Fine-Grained Image Forgery Detection and Localization},
  author    = {Guo, Xiao and Liu, Xiaohong and Ren, Zhiyuan and Grosz, Steven and Masi, Iacopo and Liu, Xiaoming},
  booktitle = {IEEE/CVF Conference on Computer Vision and Pattern Recognition (CVPR)},
  pages     = {3155--3165},
  year      = {2023},
  doi       = {10.1109/CVPR52729.2023.00308}
}

@inproceedings{yu2024diffforensics,
  title     = {DiffForensics: Leveraging Diffusion Prior to Image Forgery Detection and Localization},
  author    = {Yu, Zeqin and Ni, Jiangqun and Lin, Yuzhen and Deng, Haoyi and Li, Bin},
  booktitle = {IEEE/CVF Conference on Computer Vision and Pattern Recognition (CVPR)},
  pages     = {12765--12774},
  year      = {2024},
  doi       = {10.1109/CVPR52733.2024.01213}
}

@inproceedings{li2024unionformer,
  title     = {UnionFormer: Unified-Learning Transformer with Multi-View Representation for Image Manipulation Detection and Localization},
  author    = {Li, Shuaibo and Ma, Wei and Guo, Jianwei and Xu, Shibiao and Li, Benchong and Zhang, Xiaopeng},
  booktitle = {IEEE/CVF Conference on Computer Vision and Pattern Recognition (CVPR)},
  pages     = {12523--12533},
  year      = {2024},
  doi       = {10.1109/CVPR52733.2024.01190}
}

@inproceedings{ruiz2023dreambooth,
  title     = {DreamBooth: Fine Tuning Text-to-Image Diffusion Models for Subject-Driven Generation},
  author    = {Ruiz, Nataniel and Li, Yuanzhen and Jampani, Varun and Pritch, Yael and Rubinstein, Michael and Aberman, Kfir},
  booktitle = {IEEE/CVF Conference on Computer Vision and Pattern Recognition (CVPR)},
  pages     = {22500--22510},
  year      = {2023},
  doi       = {10.1109/CVPR52729.2023.02155}
}

@article{dong2022023mvssnet,
  author    = {Dong, Cheng and Liu, Jiantao and Zhu, Jieming and Sun, Xudong and Xu, Yongjian},
  title     = {MVSS-Net: Multi-View Multi-Scale Supervision Network for Image Manipulation Detection},
  journal   = {IEEE Transactions on Pattern Analysis and Machine Intelligence (TPAMI)},
  volume    = {45},
  number    = {3},
  pages     = {3348--3363},
  year      = {2023},
  url       = {https://ieeexplore.ieee.org/document/9876734}
}

@article{cozzolino2020noiseprint,
  author    = {Cozzolino, Davide and Verdoliva, Luisa},
  title     = {Noiseprint: A CNN-Based Camera Model Fingerprint},
  journal   = {IEEE Transactions on Information Forensics and Security (TIFS)},
  volume    = {15},
  pages     = {144--159},
  year      = {2020},
  url       = {https://ieeexplore.ieee.org/document/9022397}
}

@inproceedings{saharia2022imagen,
  author    = {Chitwan Saharia and
               William Chan and
               Saurabh Saxena and
               Lala Li and
               Jay Whang and
               Emily L. Denton and
               Seyed Kamyar Seyed Ghasemipour and
               Raphael Gontijo Lopes and
               Burcu Karagol Ayan and
               Tim Salimans and
               Jonathan Ho and
               David J. Fleet and
               Mohammad Norouzi},
  title     = {Photorealistic Text-to-Image Diffusion Models with Deep Language Understanding},
  booktitle = {Advances in Neural Information Processing Systems (NeurIPS)},
  year      = {2022},
  url       = {https://papers.nips.cc/paper_files/paper/2022/hash/ec795aeadae0b7d230fa35cbaf04c041-Abstract-Conference.html}
}

@article{ma2023imlvit,
  title   = {IML-ViT: Benchmarking Image Manipulation Localization by Vision Transformer},
  author  = {Ma, Xiaochen and Du, Bo and Jiang, Zhuohang and Hammadi, Ahmed Y Al and Zhou, Jizhe},
  journal = {arXiv preprint arXiv:2307.14863},
  year    = {2023}
}

@inproceedings{ZhangLYXLZ24,
  author       = {Xuanyu Zhang and Runyi Li and Jiwen Yu and Youmin Xu and Weiqi Li and Jian Zhang},
  title        = {EditGuard: Versatile Image Watermarking for Tamper Localization and Copyright Protection},
  booktitle    = {IEEE/CVF Conference on Computer Vision and Pattern Recognition (CVPR)},
  year         = {2024},
  pages        = {11964--11974},
  publisher    = {IEEE},
  doi          = {10.1109/CVPR52733.2024.01137},
  url          = {https://doi.org/10.1109/CVPR52733.2024.01137}
}

@inproceedings{ChuXWZYZ25,
  author       = {Beilin Chu and Xuan Xu and Xin Wang and Yufei Zhang and Weike You and Linna Zhou},
  title        = {FIRE: Robust Detection of Diffusion-Generated Images via Frequency-Guided Reconstruction Error},
  booktitle    = {IEEE/CVF Conference on Computer Vision and Pattern Recognition (CVPR)},
  year         = {2025},
  pages        = {12830--12839},
  publisher    = {IEEE},
  doi          = {10.1109/CVPR52734.2025.01197},
  url          = {https://doi.org/10.1109/CVPR52734.2025.01197}
}

@inproceedings{RickerLF24,
  author       = {Jonas Ricker and Denis Lukovnikov and Asja Fischer},
  title        = {AEROBLADE: Training-Free Detection of Latent Diffusion Images Using Autoencoder Reconstruction Error},
  booktitle    = {IEEE/CVF Conference on Computer Vision and Pattern Recognition (CVPR)},
  year         = {2024},
  pages        = {9130--9140},
  publisher    = {IEEE},
  doi          = {10.1109/CVPR52733.2024.00872},
  url          = {https://doi.org/10.1109/CVPR52733.2024.00872}
}

@article{zhou2025aigi,
  title        = {AIGI-Holmes: Towards Explainable and Generalizable AI-Generated Image Detection via Multimodal Large Language Models},
  author       = {Zhou, Ziyin and Luo, Yunpeng and Wu, Yuanchen and Sun, Ke and Ji, Jiayi and Yan, Ke and Ding, Shouhong and Sun, Xiaoshuai and Wu, Yunsheng and Ji, Rongrong},
  journal      = {arXiv preprint arXiv:2507.02664},
  year         = {2025},
  doi          = {10.48550/ARXIV.2507.02664}
}

@inproceedings{wang2023dire,
  title     = {DIRE for Diffusion-Generated Image Detection},
  author    = {Wang, Zhendong and Bao, Jianmin and Zhou, Wengang and Wang, Weilun and Hu, Hezhen and Chen, Hong and Li, Houqiang},
  booktitle = {IEEE/CVF International Conference on Computer Vision (ICCV)},
  pages     = {22388--22398},
  year      = {2023},
  doi       = {10.1109/ICCV51070.2023.02051}
}

@article{luo2024lare2,
  title   = {LaRE\textasciicircum 2: Latent Reconstruction Error Based Method for Diffusion-Generated Image Detection},
  author  = {Luo, Yunpeng and Du, Junlong and Yan, Ke and Ding, Shouhong},
  journal = {arXiv preprint arXiv:2403.17465},
  year    = {2024},
  doi     = {10.48550/ARXIV.2403.17465}
}

@inproceedings{zhu2023genimage,
  title     = {GenImage: A Million-Scale Benchmark for Detecting AI-Generated Image},
  author    = {Zhu, Mingjian and Chen, Hanting and Yan, Qiangyu and Huang, Xudong and Lin, Guanyu and Li, Wei and Tu, Zhijun and Hu, Hailin and Hu, Jie and Wang, Yunhe},
  booktitle = {Advances in Neural Information Processing Systems (NeurIPS)},
  year      = {2023},
  url       = {https://papers.nips.cc/paper_files/paper/2023/hash/f4d4a021f9051a6c18183b059117e8b5-Abstract-Datasets_and_Benchmarks.html}
}

@inproceedings{zhang2024imd_inr,
  title     = {Image Manipulation Detection with Implicit Neural Representation and Limited Supervision},
  author    = {Zhang, Zhenfei and Li, Mingyang and Li, Xin and Chang, Ming-Ching and Hsieh, Jun-Wei},
  booktitle = {European Conference on Computer Vision (ECCV)},
  pages     = {255--273},
  year      = {2024},
  doi       = {}
}

@inproceedings{li2024napl,
  title     = {Noise-Assisted Prompt Learning for Image Forgery Detection and Localization},
  author    = {Li, Dong and Zhu, Jiaying and Fu, Xueyang and Guo, Xun and Liu, Yidi and Yang, Gang and Liu, Jiawei and Zha, Zheng-Jun},
  booktitle = {European Conference on Computer Vision (ECCV)},
  pages     = {18--36},
  year      = {2024},
  doi       = {10.1007/978-3-031-73247-8_2}
}

@inproceedings{asnani2022proactive,
  title     = {Proactive Image Manipulation Detection},
  author    = {Asnani, Vishal and Yin, Xi and Hassner, Tal and Liu, Sijia and Liu, Xiaoming},
  booktitle = {IEEE/CVF Conference on Computer Vision and Pattern Recognition (CVPR)},
  pages     = {15365--15374},
  year      = {2022},
  doi       = {10.1109/CVPR52688.2022.01495}
}

@inproceedings{asnani2023malp,
  title     = {MaLP: Manipulation Localization Using a Proactive Scheme},
  author    = {Asnani, Vishal and Yin, Xi and Hassner, Tal and Liu, Xiaoming},
  booktitle = {IEEE/CVF Conference on Computer Vision and Pattern Recognition (CVPR)},
  pages     = {12343--12352},
  year      = {2023},
  doi       = {10.1109/CVPR52729.2023.01188}
}

@inproceedings{ojha2023universal,
  title     = {Towards Universal Fake Image Detectors that Generalize Across Generative Models},
  author    = {Ojha, Utkarsh and Li, Yuheng and Lee, Yong Jae},
  booktitle = {IEEE/CVF Conference on Computer Vision and Pattern Recognition (CVPR)},
  pages     = {24480--24489},
  year      = {2023},
  doi       = {10.1109/CVPR52729.2023.02345}
}

@inproceedings{chen2024drct,
  title     = {DRCT: Diffusion Reconstruction Contrastive Training towards Universal Detection of Diffusion Generated Images},
  author    = {Chen, Baoying and Zeng, Jishen and Yang, Jianquan and Yang, Rui},
  booktitle = {International Conference on Machine Learning (ICML)},
  year      = {2024},
  url       = {https://openreview.net/forum?id=oRLwyayrh1}
}

@inproceedings{cazenavette2024fakeinversion,
  title     = {FakeInversion: Learning to Detect Images from Unseen Text-to-Image Models by Inverting Stable Diffusion},
  author    = {Cazenavette, George and Sud, Avneesh and Leung, Thomas and Usman, Ben},
  booktitle = {IEEE/CVF Conference on Computer Vision and Pattern Recognition (CVPR)},
  pages     = {10759--10769},
  year      = {2024},
  doi       = {10.1109/CVPR52733.2024.01023}
}

@inproceedings{liu2023evp,
  title     = {Explicit Visual Prompting for Low-Level Structure Segmentations},
  author    = {Liu, Weihuang and Shen, Xi and Pun, Chi-Man and Cun, Xiaodong},
  booktitle = {IEEE/CVF Conference on Computer Vision and Pattern Recognition (CVPR)},
  pages     = {19434--19445},
  year      = {2023},
  doi       = {10.1109/CVPR52729.2023.01862}
}

@inproceedings{HeCXLDG22,
  author       = {Kaiming He and Xinlei Chen and Saining Xie and Yanghao Li and Piotr Dollár and Ross B. Girshick},
  title        = {Masked Autoencoders Are Scalable Vision Learners},
  booktitle    = {IEEE/CVF Conference on Computer Vision and Pattern Recognition (CVPR)},
  year         = {2022},
  pages        = {15979--15988},
  publisher    = {IEEE},
  doi          = {10.1109/CVPR52688.2022.01553},
  url          = {https://doi.org/10.1109/CVPR52688.2022.01553}
}

@article{WangHAZ25,
  author       = {Yi Wang and Hugo Hernández Hernández and Conrad M. Albrecht and Xiao Xiang Zhu},
  title        = {Feature Guided Masked Autoencoder for Self-Supervised Learning in Remote Sensing},
  journal      = {IEEE Journal of Selected Topics in Applied Earth Observations and Remote Sensing},
  year         = {2025},
  volume       = {18},
  pages        = {321--336},
  doi          = {10.1109/JSTARS.2024.3493237},
  url          = {https://doi.org/10.1109/JSTARS.2024.3493237}
}

@inproceedings{XieWYAAL21,
  author       = {Enze Xie and Wenhai Wang and Zhiding Yu and Anima Anandkumar and José M. Álvarez and Ping Luo},
  title        = {SegFormer: Simple and Efficient Design for Semantic Segmentation with Transformers},
  booktitle    = {Advances in Neural Information Processing Systems (NeurIPS)},
  year         = {2021},
  pages        = {12077--12090},
  url          = {https://proceedings.neurips.cc/paper/2021/hash/64f1f27bf1b4ec22924fd0acb550c235-Abstract.html}
}

@inproceedings{RonnebergerFB15,
  author       = {Olaf Ronneberger and Philipp Fischer and Thomas Brox},
  title        = {U-Net: Convolutional Networks for Biomedical Image Segmentation},
  booktitle    = {Medical Image Computing and Computer-Assisted Intervention - MICCAI 2015},
  year         = {2015},
  pages        = {234--241},
  publisher    = {Springer},
  series       = {Lecture Notes in Computer Science},
  volume       = {9351},
  doi          = {10.1007/978-3-319-24574-4_28},
  url          = {https://doi.org/10.1007/978-3-319-24574-4_28}
}

@inproceedings{Cao0YCDY22,
  author       = {Junyi Cao and Chao Ma and Taiping Yao and Shen Chen and Shouhong Ding and Xiaokang Yang},
  title        = {End-to-End Reconstruction-Classification Learning for Face Forgery Detection},
  booktitle    = {IEEE/CVF Conference on Computer Vision and Pattern Recognition (CVPR)},
  year         = {2022},
  pages        = {4103--4112},
  publisher    = {IEEE},
  doi          = {10.1109/CVPR52688.2022.00408},
  url          = {https://doi.org/10.1109/CVPR52688.2022.00408}
}

@article{LiuSC23,
  author       = {Weihuang Liu and Xi Shen and Chi-Man Pun and Xiaodong Cun},
  title        = {Explicit Visual Prompting for Universal Foreground Segmentations},
  journal      = {CoRR},
  year         = {2023},
  volume       = {abs/2305.18476},
  doi          = {10.48550/ARXIV.2305.18476},
  url          = {https://doi.org/10.48550/arXiv.2305.18476},
  eprint       = {2305.18476}
}

@inproceedings{NiloyBW23,
  author       = {Fahim Faisal Niloy and Kishor Kumar Bhaumik and Simon S. Woo},
  title        = {CFL-Net: Image Forgery Localization Using Contrastive Learning},
  booktitle    = {IEEE/CVF Winter Conference on Applications of Computer Vision (WACV)},
  year         = {2023},
  pages        = {4631--4640},
  doi          = {10.1109/WACV56688.2023.00462},
  url          = {https://doi.org/10.1109/WACV56688.2023.00462}
}

@article{RenHNXZW24,
  author       = {Ruyong Ren and Qixian Hao and Shaozhang Niu and Keyang Xiong and Jiwei Zhang and Maosen Wang},
  title        = {MFI-Net: Multi-Feature Fusion Identification Networks for Artificial Intelligence Manipulation},
  journal      = {IEEE Transactions on Circuits and Systems for Video Technology},
  year         = {2024},
  volume       = {34},
  number       = {2},
  pages        = {1266--1280},
  doi          = {10.1109/TCSVT.2023.3289171},
  url          = {https://doi.org/10.1109/TCSVT.2023.3289171}
}

@inproceedings{DongWT13,
  author       = {Jing Dong and Wei Wang and Tieniu Tan},
  title        = {CASIA Image Tampering Detection Evaluation Database},
  booktitle    = {IEEE China Summit and International Conference on Signal and Information Processing (ChinaSIP)},
  year         = {2013},
  pages        = {422--426},
  publisher    = {IEEE},
  doi          = {10.1109/CHINASIP.2013.6625374},
  url          = {https://doi.org/10.1109/ChinaSIP.2013.6625374}
}

@inproceedings{LinMBHPRDZ14,
  author       = {Tsung-Yi Lin and Michael Maire and Serge J. Belongie and James Hays and Pietro Perona and Deva Ramanan and Piotr Dollár and C. Lawrence Zitnick},
  title        = {Microsoft COCO: Common Objects in Context},
  booktitle    = {Computer Vision - ECCV 2014 - 13th European Conference},
  year         = {2014},
  pages        = {740--755},
  publisher    = {Springer},
  series       = {Lecture Notes in Computer Science},
  volume       = {8693},
  doi          = {10.1007/978-3-319-10602-1_48},
  url          = {https://doi.org/10.1007/978-3-319-10602-1_48}
}

@inproceedings{guan2019mfc,
  author    = {Haiying Guan and Mark Kozak and Eric Robertson and Yooyoung Lee and
               Amy N. Yates and Andrew Delgado and Daniel Zhou and
               Timoth{\'{e}}e Kheyrkhah and Jeff Smith and Jonathan G. Fiscus},
  title     = {MFC Datasets: Large-Scale Benchmark Datasets for Media Forensic Challenge Evaluation},
  booktitle = {Proc. IEEE Winter Conf. Appl. Comput. Vis. Workshops (WACVW)},
  pages     = {63--72},
  year      = {2019},
  doi       = {10.1109/WACVW.2019.00018}
}

@inproceedings{NicholDRSMMSC22,
  author       = {Alexander Quinn Nichol and Prafulla Dhariwal and Aditya Ramesh and Pranav Shyam and Pamela Mishkin and Bob McGrew and Ilya Sutskever and Mark Chen},
  title        = {GLIDE: Towards Photorealistic Image Generation and Editing with Text-Guided Diffusion Models},
  booktitle    = {International Conference on Machine Learning (ICML)},
  year         = {2022},
  pages        = {16784--16804},
  publisher    = {PMLR},
  series       = {Proceedings of Machine Learning Research},
  volume       = {162},
  url          = {https://proceedings.mlr.press/v162/nichol22a.html}
}

@inproceedings{NovozamskyMS20,
  author       = {Adam Novozámský and Babak Mahdian and Stanislav Saic},
  title        = {IMD2020: A Large-Scale Annotated Dataset Tailored for Detecting Manipulated Images},
  booktitle    = {IEEE Winter Applications of Computer Vision Workshops (WACV)},
  year         = {2020},
  pages        = {71--80},
  publisher    = {IEEE},
  doi          = {10.1109/WACVW50321.2020.9096940},
  url          = {https://doi.org/10.1109/WACVW50321.2020.9096940}
}

@inproceedings{WangHH25,
  author       = {Yabin Wang and Zhiwu Huang and Xiaopeng Hong},
  title        = {OpenSDI: Spotting Diffusion-Generated Images in the Open World},
  booktitle    = {IEEE/CVF Conference on Computer Vision and Pattern Recognition (CVPR)},
  year         = {2025},
  pages        = {4291--4301},
  publisher    = {Computer Vision Foundation / IEEE},
  doi          = {10.1109/CVPR52734.2025.00405},
  url          = {https://openaccess.thecvf.com/content/CVPR2025/html/Wang_OpenSDI_Spotting_Diffusion-Generated_Images_in_the_Open_World_CVPR_2025_paper.html}
}

@inproceedings{WangNH025,
  author       = {Kai Wang and Shaozhang Niu and Qixian Hao and Jiwei Zhang},
  title        = {InpDiffusion: Image Inpainting Localization via Conditional Diffusion Models},
  booktitle    = {AAAI Conference on Artificial Intelligence (AAAI)},
  year         = {2025},
  pages        = {7771--7779},
  publisher    = {AAAI Press},
  doi          = {10.1609/AAAI.V39I7.32837},
  url          = {https://doi.org/10.1609/aaai.v39i7.32837}
}

@inproceedings{HaoN0W25,
  author       = {Qixian Hao and Shaozhang Niu and Jiwei Zhang and Kai Wang},
  title        = {DcDsDiff: Dual-Conditional and Dual-Stream Diffusion Model for Generative Image Tampering Localization},
  booktitle    = {Proceedings of the Thirty-Fourth International Joint Conference on Artificial Intelligence (IJCAI)},
  year         = {2025},
  pages        = {1071--1079},
  publisher    = {ijcai.org},
  doi          = {10.24963/IJCAI.2025/120},
  url          = {https://doi.org/10.24963/ijcai.2025/120}
}

@inproceedings{GuillaroCSDV23,
  author       = {Fabrizio Guillaro and Davide Cozzolino and Avneesh Sud and Nicholas Dufour and Luisa Verdoliva},
  title        = {TruFor: Leveraging All-Round Clues for Trustworthy Image Forgery Detection and Localization},
  booktitle    = {IEEE/CVF Conference on Computer Vision and Pattern Recognition (CVPR)},
  year         = {2023},
  pages        = {20606--20615},
  publisher    = {IEEE},
  doi          = {10.1109/CVPR52729.2023.01974},
  url          = {https://doi.org/10.1109/CVPR52729.2023.01974}
}

@article{LiuLCL22,
  author       = {Xiaohong Liu and Yaojie Liu and Jun Chen and Xiaoming Liu},
  title        = {PSCC-Net: Progressive Spatio-Channel Correlation Network for Image Manipulation Detection and Localization},
  journal      = {IEEE Transactions on Circuits and Systems for Video Technology},
  year         = {2022},
  volume       = {32},
  number       = {11},
  pages        = {7505--7517},
  doi          = {10.1109/TCSVT.2022.3189545},
  url          = {https://doi.org/10.1109/TCSVT.2022.3189545}
}

@article{DongCHCL23,
  author       = {Chengbo Dong and Xinru Chen and Ruohan Hu and Juan Cao and Xirong Li},
  title        = {MVSS-Net: Multi-View Multi-Scale Supervised Networks for Image Manipulation Detection},
  journal      = {IEEE Transactions on Pattern Analysis and Machine Intelligence (TPAMI)},
  year         = {2023},
  volume       = {45},
  number       = {3},
  pages        = {3539--3553},
  doi          = {10.1109/TPAMI.2022.3180556},
  url          = {https://doi.org/10.1109/TPAMI.2022.3180556}
}

@inproceedings{HuZJCYN20,
  author       = {Xuefeng Hu and Zhihan Zhang and Zhenye Jiang and Syomantak Chaudhuri and Zhenheng Yang and Ram Nevatia},
  title        = {SPAN: Spatial Pyramid Attention Network for Image Manipulation Localization},
  booktitle    = {Computer Vision - ECCV 2020 - 16th European Conference},
  year         = {2020},
  pages        = {312--328},
  publisher    = {Springer},
  series       = {Lecture Notes in Computer Science},
  volume       = {12366},
  doi          = {10.1007/978-3-030-58589-1_19},
  url          = {https://doi.org/10.1007/978-3-030-58589-1_19}
}

@inproceedings{DongBZZCY00YC0Y23,
  author       = {Xiaoyi Dong and Jianmin Bao and Yinglin Zheng and Ting Zhang and Dongdong Chen and Hao Yang and Ming Zeng and Weiming Zhang and Lu Yuan and Dong Chen and Fang Wen and Nenghai Yu},
  title        = {MaskCLIP: Masked Self-Distillation Advances Contrastive Language-Image Pretraining},
  booktitle    = {IEEE/CVF Conference on Computer Vision and Pattern Recognition (CVPR)},
  year         = {2023},
  pages        = {10995--11005},
  publisher    = {IEEE},
  doi          = {10.1109/CVPR52729.2023.01058},
  url          = {https://doi.org/10.1109/CVPR52729.2023.01058}
}
}


\end{document}